\documentclass{article}
\usepackage{graphicx}
\usepackage{subfig}
\pdfoutput=1
\usepackage{svg}
\usepackage{multicol}
\usepackage{hyperref}
\usepackage{graphicx}
 \usepackage{float} 
 \usepackage{amsmath}
\usepackage{multirow}
\usepackage{algorithm,algpseudocode}
\usepackage{amssymb}
\usepackage{diagbox}
\usepackage{float}
\usepackage{pgfplots}
\usepackage{pgfplotstable}

\usepackage{filecontents}
\usepgfplotslibrary{external} 
\title{\large \bf Greybox fuzzing as a contextual bandits problem}
\author{
  Ketan Patil\\
  \texttt{patilp@iisc.ac.in}
  \and
  Aditya Kanade\\
  \texttt{kanade@iisc.ac.in}
}
\date{June 10, 2018 }

\begin{document}

\pgfplotsset{compat=1.13}
	\maketitle
    
        \vskip 12pt
	\thispagestyle{empty}
	
	\bibliographystyle{unsrt}
	
		\begin{abstract}
		Greybox fuzzing is one of the most useful and effective techniques for the bug detection in large scale application programs. It uses minimal amount of instrumentation. American Fuzzy Lop (AFL) is a popular coverage based evolutionary greybox fuzzing tool. AFL performs extremely well in fuzz testing large applications and finding critical vulnerabilities, but AFL involves a lot of heuristics while deciding the favored test case(s), skipping test cases during fuzzing, assigning fuzzing iterations to test case(s). In this work, we aim at replacing the heuristics the AFL uses while assigning the fuzzing iterations to a test case during the random fuzzing. We formalize this problem as a `contextual bandit problem' and we propose an algorithm to solve this problem. We have implemented our approach on top of the AFL. We modify the AFL's heuristics with our learned model through the policy gradient method. Our learning algorithm selects the multiplier of the number of fuzzing iterations to be assigned to a test case during random fuzzing, given a fixed length substring of the test case to be fuzzed. We fuzz the substring with this new energy value and continuously updates the policy based upon the interesting test cases it produces on fuzzing. 

	\end{abstract}

	\begin{multicols}{2}
	\section{INTRODUCTION}
	Fuzz testing is being extensively used since the last 2 decades in both the academia and industry\cite{micorsoftRisk,Godefroid2008,libFuzzer, ossFuzz}, for finding out bugs and security vulnerabilities in large scale application programs. In fuzzing, the target program is continuously executed with the provided or newly generated test cases. These test cases are generated by the genetic operations like mutation, crossover or through some symbolic execution and constraint solving operations. Depending upon the amount of exposure of the program structure of the target program, the fuzz testing can be broadly classified into 3 types, namely white box fuzz testing \cite{Godefroid2008}, grey box fuzz testing \cite{Lemieux} and black box fuzz testing \cite{BLACK}. In black box fuzz testing, no program analysis is performed and the fuzz are generated at random by mutating the bits and bytes of the provided seeds and the generated fuzz. Thus most of the times, black box fuzzers require a lot of time for creating the valid and malicious inputs which can expose the software vulnerabilities. High throughput and very low overhead are some of the advantages of the black box fuzzing. The black box fuzzing can be made more efficient by using techniques like good quality seed selections\cite{householder2012probability, rebert2014optimizing}, proper scheduling of mutations\cite{woo2013scheduling}. 
	
	In case of white box fuzz testing, the extensive program analysis is performed and mostly the symbolic execution technique is used. The test cases are generated by collecting the path constraints and then one by one negating and solving them with the help of constraint solvers. Symbolic execution and constraint solving do not scale to large programs having million lines of code and paths. Due to the path explosion issue, the constraints will grow up exponentially and it becomes impossible to solve for any solver and it results in imprecise results. Godefroid et al.\cite{Godefroid2008} have given an approach for efficient white box fuzz testing. They have used generational search algorithm instead of using classical depth first search approach which attempts to expand multiple constraints instead of just the last one. Even though this approach is faster than the classical white box fuzz testing, authors observed that the symbolic execution is very slow. So the white box fuzz testing approach does not look much practical. The grey box fuzz testing sits in between the black box fuzzing and the white box fuzzing. It uses minimal amount of instrumentation to gain some knowledge about the program structure. It uses feedback mechanism like coverage information to guide the fuzzing process. Some approaches combine the symbolic execution with fuzzing\cite{ognawala2017exploratory, stephens2016driller}. Taint analysis techniques have also been used along with fuzzing for improving the performance of fuzzing\cite{ganesh2009taint}. Recently, fuzzing using machine learning\cite{ godefroid2017learn, rajpal2017not} and reinforcement learning techniques\cite{bottinger2018deep} have been proposed.  
	
	AFL \cite{afl} is a popular coverage based evolutionary greybox fuzzing tool. AFL inserts only the minimal instrumentation to record the branch coverage. Thus the instrumentation overhead is very less compared to the white box fuzz testing. AFL \cite{WhitePaper} takes instrumented binary of the program to be tested and one or more sample input test case(s) which is/are generally referred to as seed(s). Then these seeds are mutated one by one to obtain the new test cases. The main aim of the AFL is to maximize the branch coverage. AFL has found a large number of critical vulnerabilities in popular libraries and tools like Mozilla Firefox, SQLite, Apple Safari, tcpdump, file, libxml2, lrzip, binutils, vlc \cite{afl} etc. In most of the cases, it starts with an empty file as a seed input and is able to create valid and malicious inputs which expose the crashes. The simplicity of AFL has made it very popular and efficient as compared to other fuzzers. AFL maintains a queue of interesting test cases and it dequeues test cases in round robin fashion. AFL can not figure out the best test case to be fuzzed next, instead it just uses the round robin scheduling and some heuristics for skipping a test case while fuzzing. It performs the deterministic fuzzing, random fuzzing and splicing on the currently selected test case. As a result of these operations, new test cases are generated and if they cover some new branches then they are considered as interesting test cases and are added to the queue. The AFL periodically performs the culling queue operation which marks some of the test cases as favored and and gives them more preference during all fuzzing steps.   
	 
	 In this work, we aim at replacing the heuristics the AFL uses while assigning the fuzzing iterations to a test case during the random fuzzing. AFL decides the number of random fuzzing iterations for a test case on the basis of the external features of the test case like the execution time, bitmap coverage, depth of the test case in terms of fuzzing hierarchy. But it does not consider the quality of the test case in terms of the test case contents. This may result in giving more or less number of random fuzzing iterations to a test case. So it can result in over fuzzing or under fuzzing. Some of the test cases can be pretty large (around 10KB), so it is not feasible to consider the entire test case contents while deciding the number of the random fuzzing iterations. Thus it's difficult to completely replace the heuristics with some learning mechanism. An earlier approach\cite{bohme2016coverage} discusses about the problem of assigning the energy value to the test cases based upon the density of distribution of unique paths traversed by the test cases. But they do not consider the test case contents while deciding the energy.

	 In this work, we explore whether it is possible to solve the problem of deciding the energy value from the test case contents, using reinforcement learning techniques\cite{mnih2015human}. We formalize this problem as a `contextual bandit problem'. We propose an algorithm to solve this problem. We consider the fixed length substring of the test case and treat it as a `state' in contextual bandits setting. We propose multipliers of the test case's energy and treat them as the `action space' in the contextual bandit setting. We give a `reward function' based upon the number of interesting test cases generated after fuzzing the substring with our newly derived energy value. Thus our approach takes a state as input and outputs one of the action from the action space. The action is nothing but the multiplier. We multiply the energy value predicted by the AFL for that test case with the multiplier to obtain a new energy value. We fuzz the substring with this new energy value and we calculate the rewards from the reward function and continuously update the policy through policy gradient method, using the reward obtained.

	 We have implemented our approach on top of the AFL. We modify the AFL's heuristics with our model learned through the policy gradient method. We have implemented a neural network architecture consisting of Long Short Term Memory(LSTM) to encode our state. As the state is a sequential stream of bytes so we selected LSTM for encoding. We have maintained exploration-exploitation trade-off for choosing action as well as fuzzing the entire test case vs. fuzzing the substring only. We have performed a number of experiments with various configurations of our model, across the various programs like binutils\cite{Binutils}, tcpdump\cite{tcpdump}, mpg321\cite{mpg321}, libpng\cite{libpng}, gif2png\cite{gif2png}, libxml2\cite{libxml2}. We have also performed the cross-binary experiments across the binutils binaries, where we train our model on one binary and test it on the other binaries. We compare the coverage, total number of paths and crashes resulted from these experiments.    
	 
	 In summary, we make the following contributions:
	 \begin{itemize}
     \item We formalize the problem of deciding the energy as a `contextual bandit problem'.
     \item We present an algorithm to decide the energy multiplier of a test case given a fixed length contents of the test case. 
     \item We implement our neural network based learning algorithm on top of the AFL and we compare results of different configurations of our model with AFL.
     \item We integrate AFL's code with a popular open-source machine learning framework tensorflow. This implementation can be useful for others while implementing any machine/reinforcement learning technique with fuzzers like AFL.  
     \end{itemize}
	
	\section{BACKGROUND}
	In this section, we explain AFL, Multi-armed Bandits problem and Contextual Bandits problem in detail.
	\subsection{AFL}
American Fuzzy Lop (AFL) is a popular coverage based evolutionary greybox fuzzing tool. The input given to AFL is an instrumented binary and one or more valid input tests called as seed(s). The main aim of the AFL is to maximize the branch coverage. The lightweight instrumentation for counting the branch coverage is added at the compile time in the target program. The algorithm \ref{alg:the_alg} explains the working of AFL. 

AFL first of all sets up the shared memory. AFL uses 64 KB of shared memory known as trace bits which captures the execution of individual test case and 64 KB of virgin bits which captures the overall effect of all the test cases executed so far. It initializes each byte in the trace bits to 0. Each byte represents a branch. AFL then reads the test cases which are given as seeds, and shuffle them and enqueue them in a queue. It stores information about each seed like it's length, file name, execution checksum, execution time, bitmap size, whether the test case is favored or not etc. whenever computed.

AFL executes these seeds one by one and the trace bits will get updated because of the instrumentation. It records the count of execution of each branch and then bucket them up in the buckets of different capacities (powers of 2). The virgin bits will be modified if the test case executes entirely a new branch or it executes already executed branch with the count which falls under a bucket different than the earlier occupied buckets. This helps in skipping tests which are executing repetitive branches. The different buckets also provide some degree of immunity against tuple collisions\cite{WhitePaper}. The path executed by a test case is identified uniquely from the checksum of the trace bits. AFL calculates this checksum by a hash function over the trace bits. AFL also counts the number of bytes set in the trace bits by each test case. AFL store this information with that test case in a field named bitmap size. This bitmap size proves useful while comparing the two different fuzz.

\begin{algorithm}[H]
\caption{AFL}
\label{alg:the_alg}
\begin{algorithmic}[1]
    \Function{AFL\_FUZZ}{$executable, seeds$}
    \State \textrm{Setup the shared memory}
    \State $Queue \gets seeds$
    \For{seed $\in$ $Queue$ }
	\State \textrm{Execute the $executable$ with seed}
        \State \textrm{Update the trace bits and virgin bits}
        \State \textrm{Calculate the fav\_factor}
        \State \textrm{Setup the topRated test cases}
                                                
    \EndFor
    \State \textrm{Perform culling operation}
     \For{testcase $\in$ Queue }
        \State $energy \gets PerformanceScore(\textrm{testcase})$
        \For{$i = 1$ to $Length(testcase)$}
            \State \textrm{Mutate testcase deterministically}  
            \If{\textrm{fuzz is interesting}}
                \State \textrm{Add fuzz to queue}
            \Else
                \State \textrm{Ignore the fuzz}
            \EndIf    
        \EndFor   
        \For{$i = 1$ to $energy$}
                \State \textrm{Mutate testcase randomly}                    
                
                \If{\textrm{fuzz is interesting}}
                    \State \textrm{Add fuzz to queue}
                \Else
                    \State \textrm{Ignore the fuzz}
                \EndIf
        \EndFor
    \EndFor   
    \EndFunction
\end{algorithmic}
\end{algorithm}

Once a test case is executed the $fav\_factor$ is calculated for that test case. It is simply the product of the execution time and length of the test case. AFL maintains a top rated test case for each byte. A top rated test case for a byte means, the test case which is having the least $fav\_factor$ and executing the branch corresponding to that byte. It takes the bytes set by the test case one by one and if the current test case is having $fav\_factor$ less than the existing top rated test case, for that byte, then it sets the current test case as the top rated for that corresponding byte. Also it is required to store the count of the bytes for which the given test case is top rated. If this counter is non zero then it helps AFL in storing minimized trace of the entire trace of that test case. This minimized trace is helpful in deciding the favored test case. If a seed itself crashes or if it generates timeout then AFL can stop the further execution and may prompt user to change such seeds. It is advised by the AFL to keep these seeds small and non repetitive. AFL can also minimize these test cases while preserving their meaning. The above procedure is followed for all the seeds. If any test case is no longer top rated for any of the byte then there is no point in storing the minimized trace of that test case and wasting the space, thus AFL frees the memory allocated by those minimized trace.

Before performing any mutation on any of the seed, AFL performs culling queue operation. This operation is performed in order to give less importance to older, lengthy and less favored tests and give more importance to the more efficient test cases. It goes over each byte and checks the top rated test case for each byte, once it finds a top rated test case it checks the minimized trace of that test case and makes this test case favored for all those non zero bytes in the trace. Storing minimized trace is very useful in this case, otherwise the entire test case needs to be re-executed. The favored test cases are given more preference over non favored test cases while fuzzing.

AFL dequeues test cases one by one and fuzzes them to generate next generation fuzz. This operation is inspired from the Genetic algorithms which include the techniques like cross-overs and mutations. AFL can not figure out the best test case to be fuzzed next, instead it just uses the round robin scheduling and some heuristics for skipping a test case while fuzzing. AFL performs 2 types of fuzzing, first is deterministic fuzzing (performed only once per test case) and second one is random fuzzing. If the current test case is not favored or it is already fuzzed and there are some favored test cases in the queue which are not yet fuzzed then it skips the current test with 99$\%$ probability. It is done in order to promote shorter duration test cases which can cover more branches. Even if there is no pending favored test case and the current test case is already fuzzed then it is skipped with 95$\%$ probability. It will promote the test cases which are not yet fuzzed over the test cases which are already fuzzed at least once. But if there is no pending favored test case and the current test case is not fuzzed then the chances of skipping the test case gets reduced to 75$\%$.  

AFL then decides the number of random fuzzing iterations(also called as the energy) for that test case on the basis of the external features of the test case like the execution time, bitmap coverage, depth of the test case in terms of fuzzing hierarchy. But it does not consider the test case contents. Tests that are fast, which cover more branches and have more depth are given more energy. This energy assignment is totally heuristics based. In deterministic fuzzing, AFL perform mutations like single bit flip (flip only a single bit of the test case to generate new fuzz), two walking bits flip, four walking bits flip, byte flip, two walking bytes flip, four walking bytes flip. After each flip it executes the new fuzz formed and checks whether it is interesting or not (i.e. whether it affects virgin bits or not). If the fuzz is not found to be interesting then it is skipped and not added to the queue. It also performs 8 bit, 16 bit and 32 bit arithmetic increment and decrement operations, where it goes on selecting next word and goes on incrementing it till some max value is reached. It also sets the bytes/words/double words to interesting values like powers of 2, corner values like near to the both negative and positive overflow values. In random fuzzing it performs iterations equal to the energy given to each test case. It randomly flips a bit somewhere, sets a random byte/word/double word to some interesting value, sets random value to a random byte, randomly adds or subtracts to/from random byte, deletes random bytes, clones random bytes. It also splice the generated test cases, it selects one input file and randomly chooses other file and splices them at some random offset. It is similar to the process of producing next generation child from the existing species by using crossover operator.    
\subsection{Reinforcement Learning}
Reinforcement learning is an online learning technique used for decision making based upon the evaluative feedback rather than the instructive feedback. It has been used extensively in the study of atari games\cite{mnih2015human}, animal behavior\cite{holroyd2002neural}. The agent trained using these techniques was able to beat the human world-champion in the games like backgammon\cite{tesauro1995td}, game of Go, atari games. The markov decision process(MDP) is used by reinforcement learning to formulate the interaction between an agent and its environment in terms of states, actions and rewards. Formally, MDP M is a tuple M =($S, A, P, R$, $\gamma$) where $S$ denotes a set of states, $A$ denotes a set of actions, $P$ denotes the state transition probability matrix $P_{ss'}^{a} = \mathbb{P}(S_{t+1}= s'|S_t =s,A_{t} = a)$, $R$ denotes the reward function and $\gamma$ denotes the discount factor. In reinforcement learning, the agent starts in some state $s\in S$ , and takes an action $a\in A$ out of all the available actions from the action space. The environment gives a scalar reward $r_{t}$ to the agent and the system transits to the next state $s'$. The agent's main aim is to maximize the expected sum of discounted future reward, $G_t = \sum_{k = 0}^{T-t-1}\gamma^{k}r_{t+k}$\cite{sutton1998reinforcement}. Choosing certain action may give bad immediate rewards but it may lead to better future rewards. In contrast, choosing an action which gives better immediate reward may lead to very poor future rewards. Thus the agent has to deal with these conditions carefully, while selecting an action in a given state. Thus the action selection is decided according to the agent's policy $\pi(a|s)=\mathbb{P}(A_t=a|S_t=s)$\cite{rlCourse}.

\subsection{Multi-armed Bandits Problem}
The multi-armed bandits problem was derived from the slot machine with multiple levers. The player plays one of the levers and obtains some reward. The main aim of the player is to maximize the rewards by selecting the best action. Formally, the k-armed bandits problem consists of k actions $A =\left \{ a_1,a_2,a_3,...,a_k \right \}$. Each action $a_i$ is associated with a corresponding reward distribution $D_i$ . Given some fixed number of trials $t = 1, 2,...,n$, at each trial t, the agent selects an action $a_i \in A$ and receives a scalar reward $r_t$, where $r_t \sim D_t$. The agent does not know about the reward distributions associated with each action. The agent's goal is to maximize the expected total reward over some time period. The actual value of an action is given by $q_*(a_i)=\mathbb{E}(R_t|A_t=a_i)$ where $a_i \in A$. The agent tries to estimate this value for each action from its trials. The agent estimates the value of an action $a_i$ at time $t$ as $Q_t(a_i)$ and it's aim is to make $Q_t(a_i)$ close to $q_*(a_i)$. The agent achieves this by having a trade-off between exploration(trying out random actions) and exploitation (choosing an action which has provided the highest $q_*$ value at that time instance). Exploitation is useful in maximizing the expected rewards for a single step but the combination of exploration and exploitation is useful for achieving the higher rewards in long run. More exploration in initial phase and then gradually increasing exploitation is one of the techniques for balancing exploration and exploitation. In contrast to full reinforcement learning problem, the multi-armed bandits problem does not have a concept of episode and the rewards depend only upon the action selected. A number of solutions have been proposed to solve this multi-armed bandits problem.\cite{audibert2010best}

\subsection{Contextual Bandits Problem}
The Contextual Bandits problem falls in-between the full reinforcement learning problem and multi-armed bandits problem. Formally, contextual bandits problem consists of a state space $S = \left \{ s_1,s_2,...,s_n \right \}$ and an action space $A =\left \{ a_1,a_2,a_3,...,a_k \right \}$. At each time instance $t$, the environment presents a state $s_i \in S$ to the agent, the agent takes an action $a_j \in A$ and gets some reward $r_{ij}$ from the environment. Each state, action pair $<s_i,a_j>$ is associated with a corresponding reward distribution $D_{i,j}$. It means that the reward $r_{ij}$ is obtained from the distribution $D_{i,j}$. In case of multi-armed bandits problem, there is only a single state and multiple actions. But in case of contextual bandits problem, the reward depends upon the state as well as the action taken at that instance. The goal of the agent in this case, is not only to find out the best action for a single state, but it needs to find out the best actions for all the states, and to maximize the expected total reward over some time period. The agent improves it's policy based upon it's observations $(s,a,r)$. The contextual bandit problems are also known as associative search tasks. A variety of problems have been modeled using contextual bandits problem like personalized news recommendation system\cite{li2010contextual}, optimizing random allocation in clinical trials, adaptive routing in networks etc. The main difference between the contextual bandits problem and a full reinforcement learning problem is the state transition. In the case of full reinforcement learning problem the next state depends upon the current state and the action taken by the agent. Thus while taking an action in any state, the agent needs to take delayed rewards into the consideration. In contrast, for contextual bandits, the agent does not need to consider a long episode, rather a single state, action, reward triple forms an episode, and agent just focusses on selecting an optimal action for the given state.  

\section{Technical Details}
\subsection{Formulation}
We model the problem of deciding the number of random fuzzing iterations as a contextual-bandits problem. In this section, we describe all the components of this formulation.
\subsubsection{States}
We consider the continuous stream of 128 bytes from currently selected test case from the queue of test cases. Let us consider $Q$ denotes the queue of test cases maintained by the AFL and $T \in Q$ is the currently selected test by AFL for fuzzing. We denote $o$ as the random offset for the test case $T$, where $o \in [1, len(T)]$ where $len(T)$ denotes the length of the test case $T$. Thus our state $S$ can be represented as, 
\begin{gather*}
     S = b_{o}b_{o+1}b_{o+2}.....b_{o+127}, \\
     \enskip \textrm{where} \enskip b_{i} \enskip \textrm{represents the} \enskip i^{th} \enskip \textrm{byte in} \enskip T
\end{gather*}
If the $o+127 > len(T)$ then we append $(o+127) -len(T)$ number of zero bytes at the end of the test case $T$. In short, we take a substring of 128 contiguous bytes from the current test case and treat it as the state in the contextual bandits setting.
\subsubsection{Actions}
We define the set of multipliers as the action space. Currently our action space $A$ is,
\begin{gather*}
    A = \left \{ 0.50, 0.75, 1, 1.25, 1.50 \right \}
\end{gather*}
In order to preserve the global context of the test case we do not change the energy calculated for the test case by the AFL. After that, we decide one of the action $a \in A$ on the basis of our state $s$ and multiply the energy predicted by the AFL with the $a$, to obtain new energy value. 
\subsubsection{Rewards}\label{Rewards}
We define a reward function $R$ as follows, 
\begin{gather*}
    R = \frac{\textrm{Interesting  test cases generated}}{\textrm{total test cases generated}} 
\end{gather*}
Note that in the above formula, we are only considering the test cases generated by fuzzing our state $s$ i.e. by fuzzing a substring of 128 continuous bytes selected as state. Interesting  test cases generated = number of test cases added to the queue by AFL on fuzzing the state $s$. total test cases generated = number of test cases added to the queue by AFL + number of test cases not added to the queue by AFL(i.e. the uninteresting test cases generated by AFL on fuzzing).

\subsection{Algorithm}
We give the following two algorithms, the algorithm \ref{alg:training_alg} explains the training procedure while the other one explains the testing procedure.
\begin{algorithm}[H]
\caption{Training Algorithm}
\label{alg:training_alg}
\begin{algorithmic}[1]
    \Function{AFL\_FUZZ}{$executable, seeds$}
    \State \textrm{Setup the shared memory}
    \State $Queue \gets seeds$
    \For{seed $\in$ $Queue$ }
        \State \textrm{Execute the $executable$ with seed}
        \State \textrm{Update the trace bits and virgin bits}
        \State \textrm{Calculate the fav\_factor}
        \State \textrm{Setup the topRated test cases}
                                                
    \EndFor
    \State \textrm{Perform culling operation}
     \For{testcase $\in$ Queue }
        
        \State $energy \gets PerformanceScore(\textrm{testcase})$
        \If{$randomNumber < FuzzingProb$}
            \For{$i = 1$ to $energy$}
                \State \textrm{Mutate testcase randomly}                    
                \If{\textrm{fuzz is interesting}}
                    \State \textrm{add fuzz to queue}
                \Else
                    \State \textrm{ignore the fuzz}
                \EndIf
            \EndFor
        \Else
            \State $state \gets getState(\textrm{testcase})$ 
            \If{\textrm{random number} $< \epsilon$}
                \State $multiplier \gets$ \textrm{Random Action}
            \Else
                \State $multiplier \gets$ \textrm{Query Model} $(state)$
            \EndIf
            \State $energy \gets energy \times multiplier$
            \State $counter1, counter2 \gets 0$
            \For{$i = 1$ to $energy$}
                \State \textrm{Mutate $state$ randomly}                    
                \If{\textrm{fuzz is interesting}}
                    \State \textrm{Increment the} $counter1$
                    \State \textrm{Add fuzz to queue}
                \Else
                    \State \textrm{Ignore the fuzz}
                \EndIf
                \State \textrm{Increment the} $counter2$
            \EndFor
            \State \textrm{Calculate the reward from counters} 
            \State \textrm{Update model weights}
        \EndIf
        
    \EndFor   
    \EndFunction
\end{algorithmic}
\end{algorithm}
	 In Algorithm \ref{alg:training_alg}, we first of all initialize a variable $FuzzingProb$ with a value between 0 and 1. This probability value decides whether the entire test case will be fuzzed or just the state. 
	 \begin{table*}[t]
  \centering
   \bgroup
\def\arraystretch{1.5}
  \begin{tabular}{|c| c| c| c|c|c|c|c|c|c|c|} 
  \hline 
\multirow{2}{*}{Binary} &

 \multicolumn{3}{|c|}{Coverage} &
 \multicolumn{3}{|c|}{Crashes} &
 \multicolumn{3}{|c|}{Total Paths} \\ \cline{2-10}

  & AFL & AFL -d & AFL CB & AFL & AFL -d & AFL CB & AFL & AFL -d & AFL CB\\  
 \hline
 addr2line & 6.23 & 6.78 & 6.23 & 4 & 5 & 3 & 2427 & 3115 & 2351\\ 
 \hline
 cxxfilt & 10.15 & 10.78 & 9.36 & 532 & 1422 & 369 & 8422 & 13592 & 8198\\ 
 \hline
  elfedit & 0.18 & 0.18 & 0.18 & 0 & 0 & 0 & 21 & 21  & 21 \\ 
 \hline
  nm & 12.53 & 15.12 & 11.31 & 109 & 1179 & 44 & 4841 & 11933 & 4282 \\ 
 \hline
  objcopy & 12.53 & 14.57 & 11.98 & 30 & 94 & 33 & 3053 & 6277 & 4155\\ 
 \hline
  objdump & 10.32  & 11.69 & 10.68 & 3 & 86 & 27 & 4099 & 7198 & 5490\\ 
 \hline
  readelf & 12.43 & 15.45 & 12.99 & 0 & 0  & 0 & 5706 & 13342 & 7683 \\ 
 \hline
  size & 6.41 & 7.15 & 6.60 & 1 & 0 & 0 & 2359 & 3247 & 2452\\ 
 \hline
  strings & 0.15 & 0.15 & 0.15 & 0 & 0 & 0 & 72 & 72 & 72\\ 
 \hline
  strip-new & 11.16 & 12.98 & 11.34 & 8 & 22 & 10 & 2836 & 5860 & 4099 \\ 
 \hline
  gif2png & 1.74 & 1.76 & 1.74 & 125 & 133 & 132 & 1504 & 1599 & 1538 \\ 
 \hline
  libxml2 & 8.74 & 9.42 & 9.34 & 0 & 0 & 0 & 3211 & 5233 & 3933\\ 
 \hline
  libpng & 5.05 & 5.32 & 5.07 & 0 & 0 & 0 & 2112 & 2679 & 1957\\ 
 \hline
  mpg321 & 0.44 & 0.44  & 0.44 & 14 & 14 & 14 & 168  & 169 & 169\\ 
 \hline
  tcpdump & 15.62 & 16.36 & 17.83 & 170 & 304 & 289 & 6471 & 11018 & 8135\\ 
 \hline 
  \end{tabular}
  \egroup
  \caption{Comaprison of AFL, AFL without deterministic phase and AFL with contextual bandits}
  \label{tab:1}
\end{table*}
	 
	 The main aim of this algorithm is to predict the correct energy for the stream of 128 contiguous bytes selected from the test case, which we call as the current state. But if we fuzz only the state, then we can not perform the operations like delete and clone bytes. Thus the size of test case will never grow or reduce during fuzzing. In order to prevent such conditions, we maintain a balance between the fuzzing the entire test case and fuzzing only the state. We generate a random number between 0 and 1, if it comes out to be less than the $FuzzingProb$ then we perform random fuzzing on the complete test case with the energy decided by the AFL. Otherwise, we just perform random fuzzing on the state. In this case, we first need to decide whether we want to do exploration or exploitation. We use $\epsilon$-greedy policy to handle the trade-off between exploration and exploitation. In this policy, we explore with the probability $\epsilon$ and exploit with the probability $1-\epsilon$, in actual implementation, we again generate a random number between 0 and 1 and perform exploration if it is less than $\epsilon$, otherwise we exploit. In exploration, we choose randomly an action from the action-space. In exploitation, we query our model, which we explain in the next section, for the action to select by giving the state as an input. We multiply the energy with the action predicted, to obtain a new energy value. We perform the random fuzzing only on the state with this new energy value. We count the number of interesting test cases and total test cases generated by fuzzing the state. We calculate reward as per the reward function mentioned in the section \ref{Rewards}. We update model weights with the loss function which is calculated from the reward. Note that we are not performing the deterministic fuzzing in our approach in order to emphasize on the problem of energy decision for a state for random fuzzing. During experiment, we also disable the deterministic fuzzing from the original AFL with the help of `-d' option provided by the AFL. We can perform this training for some pre decided time period. 
	 
	 The testing algorithm is similar to the algorithm \ref{alg:training_alg}, with few changes. We don't update our trained model. Hence in testing algorithm, there is no need to count the interesting and total test cases, rewards and to update the model weights. Thus the lines 30, 34, 39, 41, 42 are not present in the testing algorithm.

\subsection{Model}	 
\begin{table*}[t]
  \centering
  \bgroup
\def\arraystretch{1.5}
\begin{tabular}{|c|c|c|c|c|c|c|c|c|c|c|}\hline
\diagbox{Training\\ Program}{Test\\ Program}&
  addr2line & cxxfit & elfedit & nm-new & objcopy & objdump & readelf & size & strings & strip-new \\ \hline
    addr2line & 6.07  & 8.57 & 0.18 & 13.30 & 12.88 & 11.05 & 14.04 & 6.88 & 0.15 & 12.70 \\ \hline
    cxxfilt & 4.89 & 9.27 & 0.16 & 13.30 & 14.06 & 11.34 & 14.76 & 7.17 & 0.15 & 12.92 \\ \hline
    elfedit & 5.87 & 9.72 & 0.18 & 13.02 & 14.97 & 11.80 & 14.22 & 7.01 & 0.15 & 13.65 \\ \hline
    nm-new & 5.64 & 9.49 & 0.18 & 6.42 & 13.44 & 11.31 & 14.44 & 6.78 & 0.15 & 13.61 \\ \hline
    objcopy & 5.91  & 10.07 & 0.18 & 10.78 & 12.13 & 11.34 & 12.77 & 7.02 & 0.15 & 12.98 \\ \hline
    objdump & 5.93 & 9.54 & 0.16 & 13.81 & 13.49 & 10.93 & 13.99 & 7.13 & 0.15 & 13.85 \\ \hline
    readelf & 5.50 & 9.76 & 0.16 & 13.19 & 13.37 & 11.34 & 11.89 & 6.92 & 0.15 & 12.09 \\ \hline
    size & 5.32  & 9.87 & 0.16 & 14.20 & 13.80 & 11.42 & 14.83 & 6.37 & 0.15 & 12.37 \\ \hline
    strings &  5.98 & 9.19 & 0.18 & 12.13 & 12.75 & 11.67 & 13.84 & 7.16 & 0.15 & 13.32 \\ \hline
    strip-new & 5.54 & 9.59 & 0.16 & 14.67 & 13.30 & 11.90 & 14.98 & 6.91 & 0.15 & 11.94 \\ \hline
    AFL -d & 6.78 & 10.78 & 0.18 & 15.12 & 14.57 & 11.69 & 15.45 & 7.15 & 0.15 & 12.98  \\ \hline
 \end{tabular}
 \egroup
  \caption{Cross-binary experiment results}
  \label{tab:2}
\end{table*}
We construct a policy based agent. As the input state is a sequential data, we use LSTM to encode our state. We use a single layer LSTM with 100 recurrent units and the default tanh activation function. Our state is a stream of 128 bytes, so we create a $128 \times 8$ matrix, where each row contains a byte represented in the binary format. We then pass it to LSTM as an input. Our model also consist of a single fully connected layer feed forward neural network with the softmax activation function and a random uniform weight initializer. This neural network takes the final hidden state of the LSTM as an input and outputs a vector of probabilities over the actions. We use the value of $\epsilon = 0.1$. We use gradient descent optimizer with learning rate = 0.001. We are using the policy gradient network to  directly update the policy, hence we use the loss function $=-log(\pi) \times R $, where $\pi$ represents the current policy i.e. the value output by out network for that corresponding action and $R$ is the reward obtained.

We have implemented our model using a popular open-source machine learning framework tensorflow-1.4.0 \cite{tensorflow}. The original AFL's code is in C and C++. There are some machine learning libraries which use C or C++ code, like tensorflow C++ API, Caffe2\cite{Caffe2}, mlpack\cite{mlpack}. But these APIs do not support all the  operations supported by the python libraries and are currently in the development phase. Hence we decided to use python for writing out model code. For integration we use ctypes\cite{ctypes}, a foreign function library for Python. We treat our AFL as an environment and we write the code for agent in python using ctypes libraries. We create a shared object of the fuzzer's code during compilation. And ctypes loads this shared object and it can call any AFL's function from the python code. ctypes also provide the C-compatible data types to make the calling of C functions and returning of the values easy and efficient. This implementation can be useful for others while integrating any machine/reinforcement learning technique with fuzzers like AFL. Our implementation is publicly available at  \url{https://bitbucket.org/iiscseal/bandit_afl}

\section{Experiments}
We used the $fuzzingProb$ value equal to 0.4. All experiments were performed on Intel(R) Xeon(R) E5-2450 2.10 GHz machine. We have performed a number of experiments with various configurations of our model, across the various popular programs and binaries like binutils-2.26, tcpdump-4.5.1, mpg321-0.3.2, libpng-1.6.32, gif2png-2.5.8, libxml2-2.9.2. 

\begin{figure*}

  \subfloat[addr2line coverage vs. time]{
\includegraphics{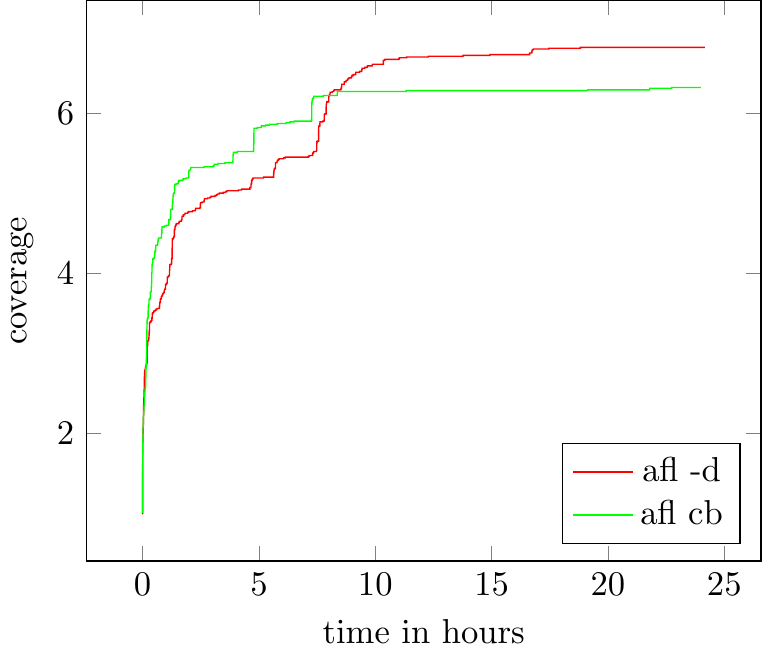}}\quad
\subfloat[cxxfilt coverage vs. time]{
\includegraphics{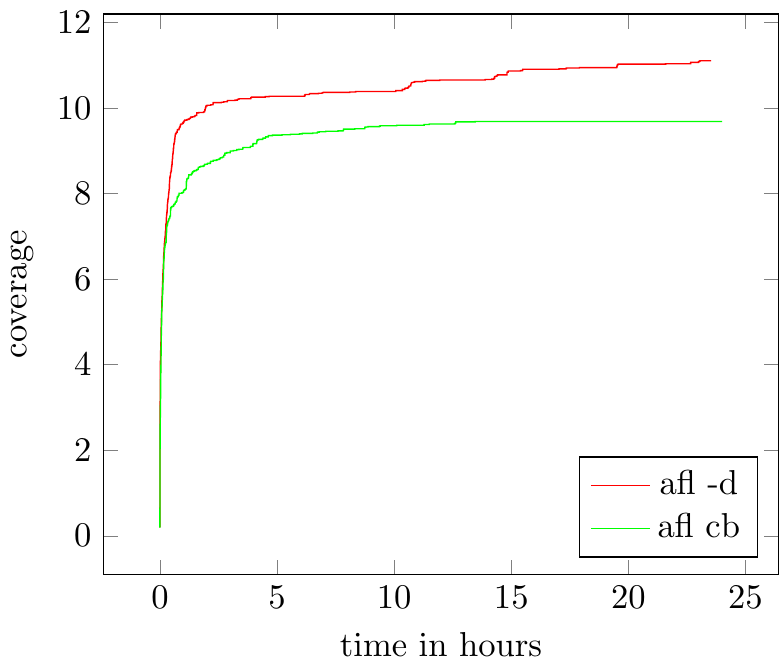}}\newline
\subfloat[nm coverage vs. time]{
\includegraphics{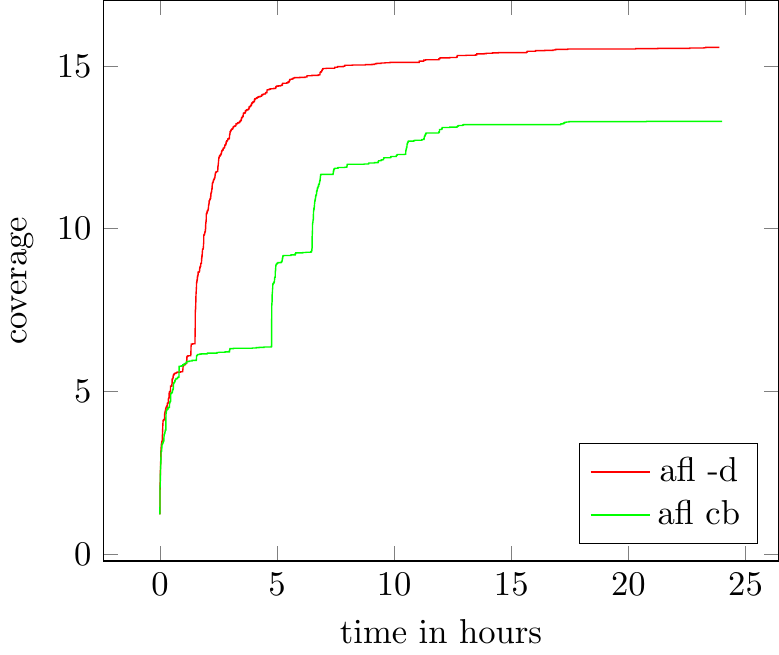}}\quad
\subfloat[objcopy coverage vs. time]{
\includegraphics{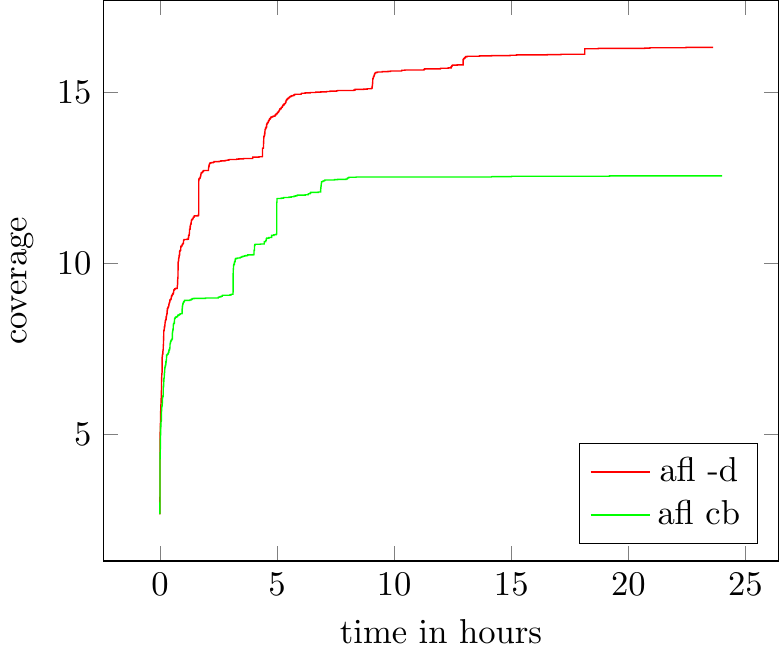}}\newline
\subfloat[objdump coverage vs. time]{
\includegraphics{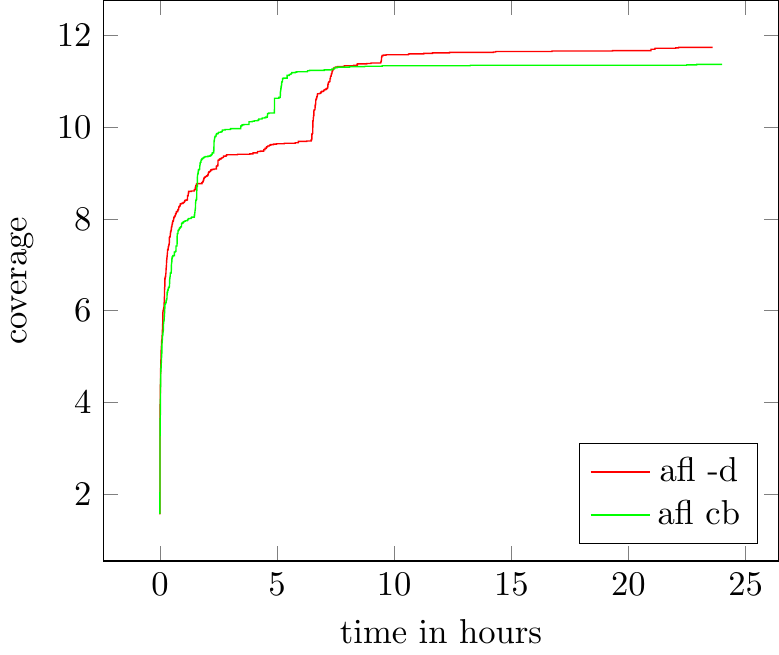}}\quad
\subfloat[readelf coverage vs. time]{
\includegraphics{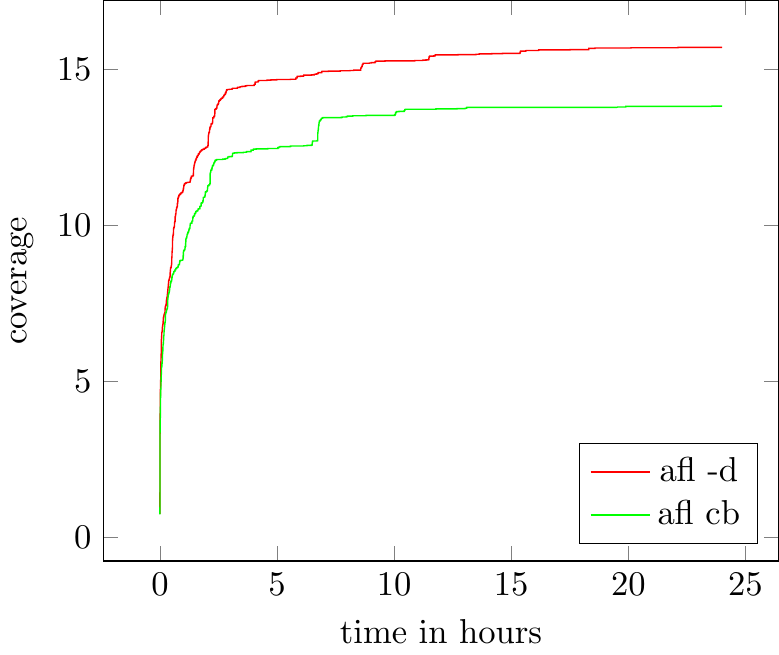}}
\end{figure*}
\begin{figure*}
\subfloat[size coverage vs. time]{
\includegraphics{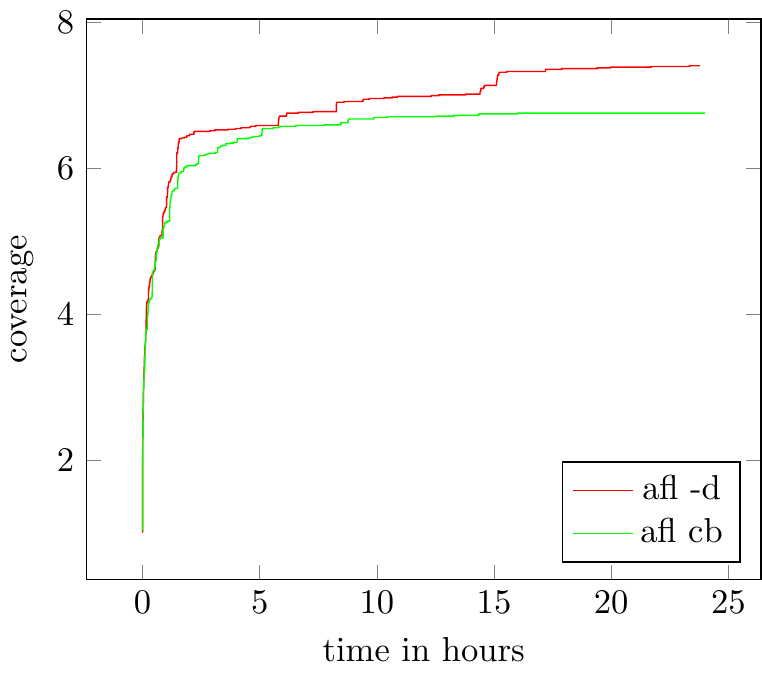}}\quad
\subfloat[srip-new coverage vs. time]{
\includegraphics{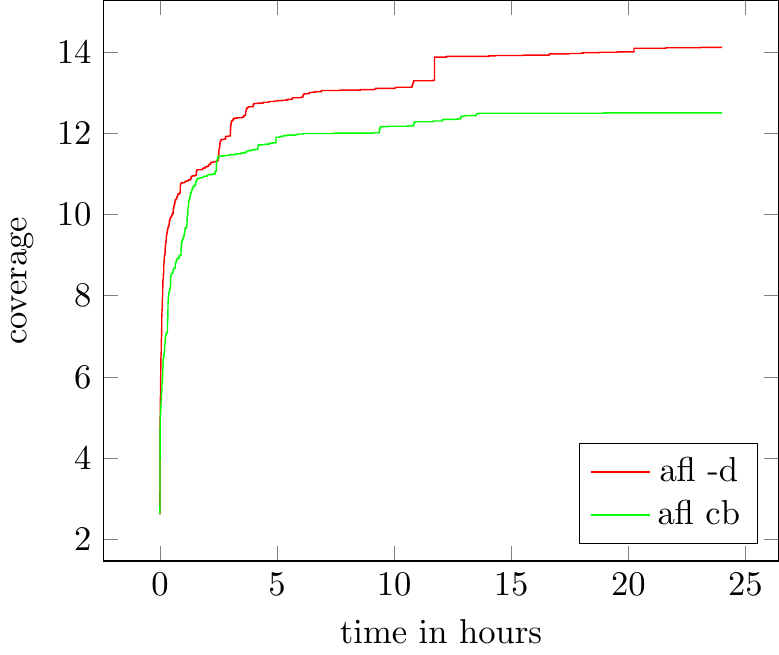}}\newline
\subfloat[gif2png coverage vs. time]{
\includegraphics{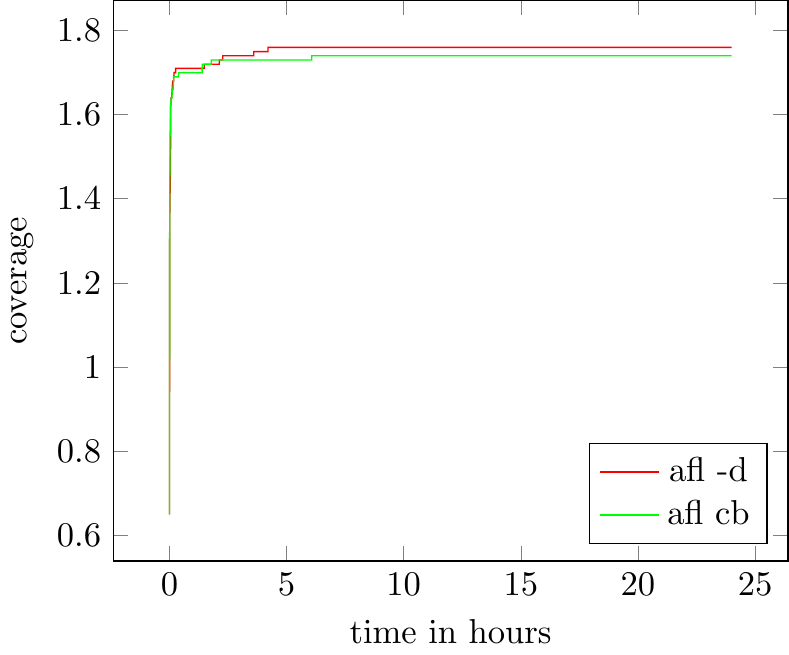}}\quad
\subfloat[libxml2 coverage vs. time]{
\includegraphics{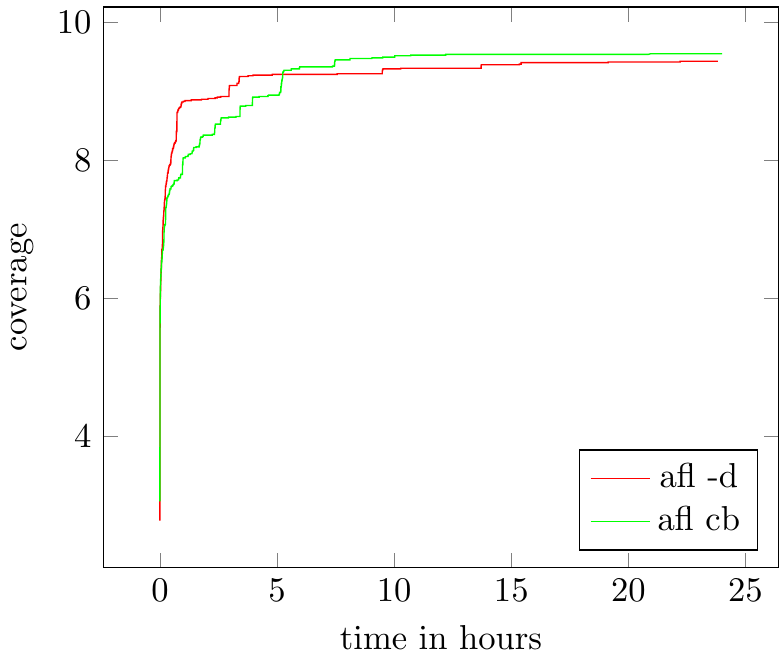}}\newline
\subfloat[libpng coverage vs. time]{
\includegraphics{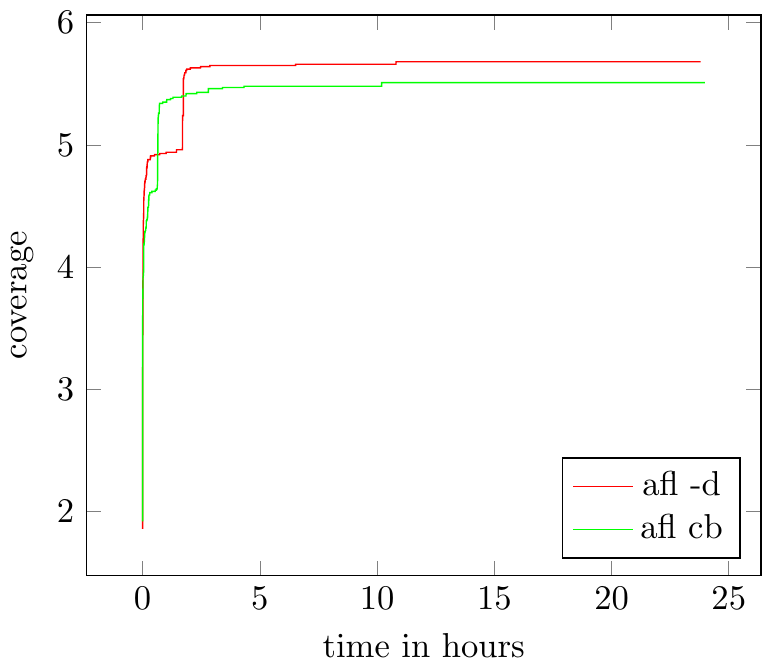}}\quad
\subfloat[tcpdump coverage vs. time]{
\includegraphics{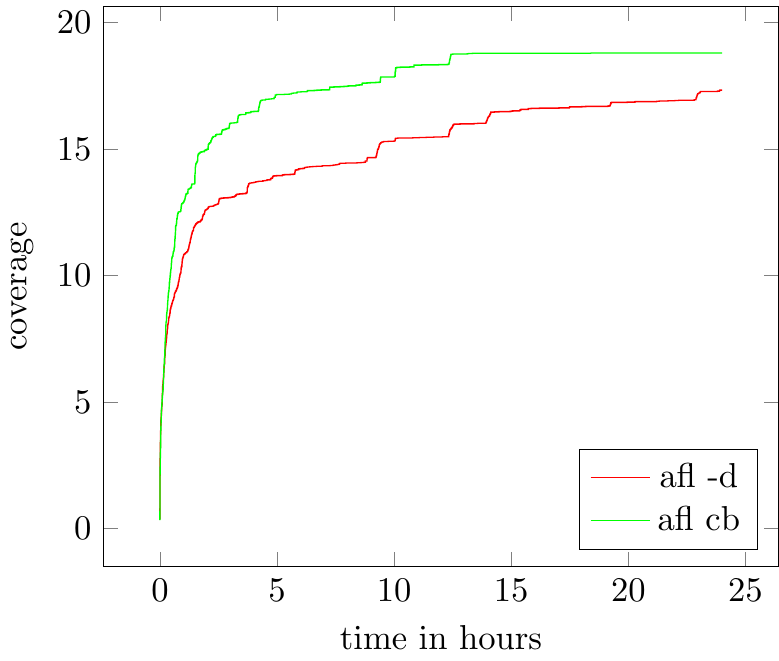}}
\end{figure*}
\begin{figure*}
\subfloat[addr2line training reward vs. iterations]{
\includegraphics{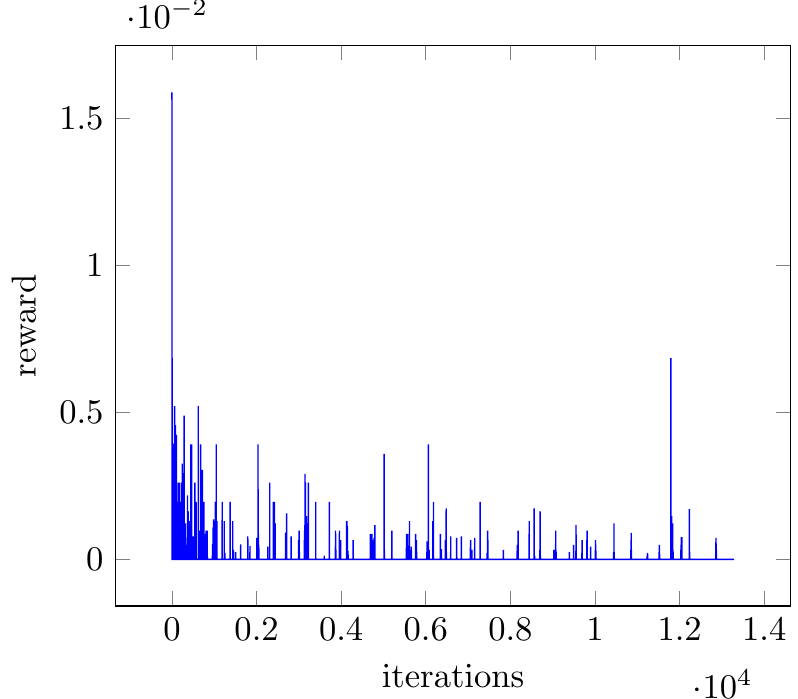}}\quad
\subfloat[cxxfilt training reward vs. iterations]{
\includegraphics{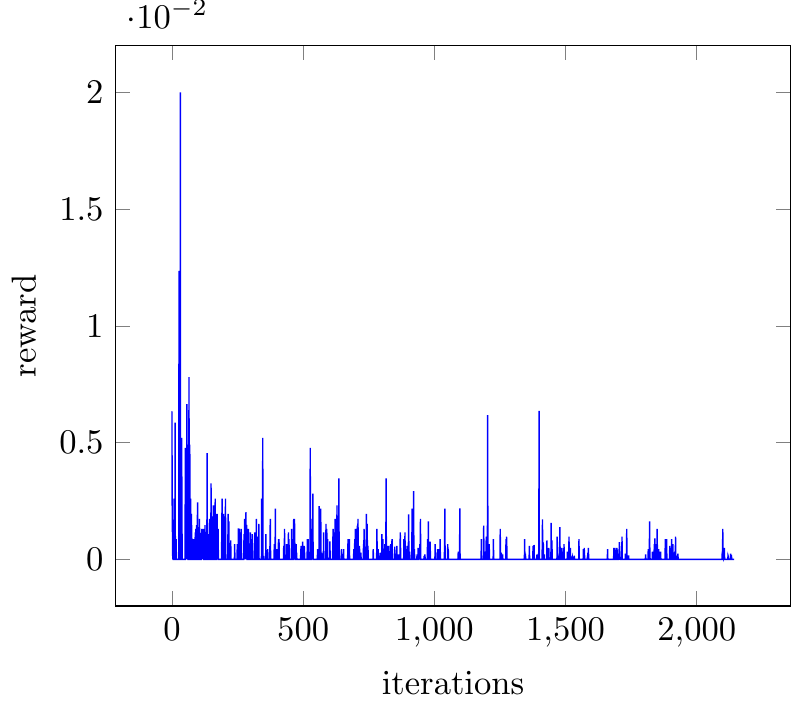}}\newline
\subfloat[nm training reward vs. iterations]{
\includegraphics{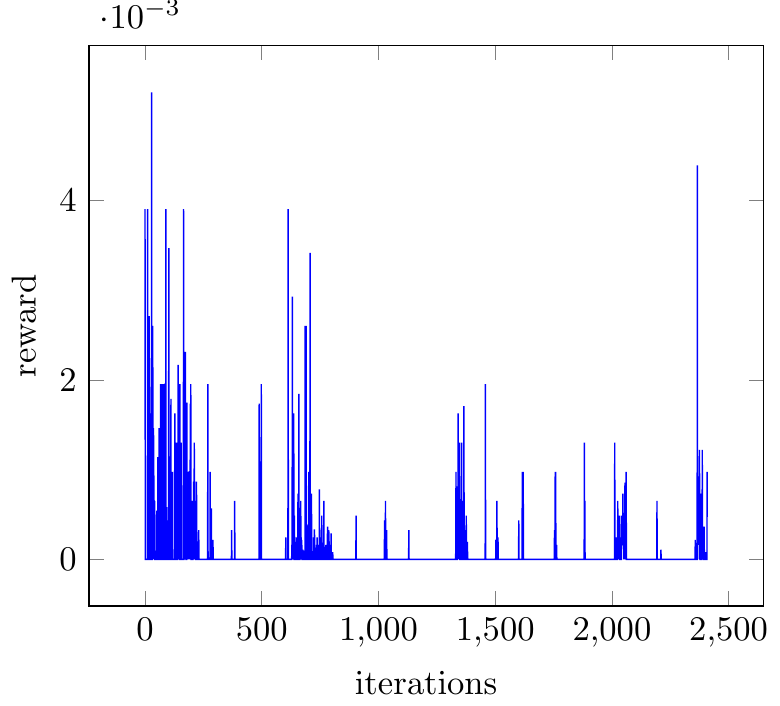}}\quad
\subfloat[objcopy training reward vs. iterations]{
\includegraphics{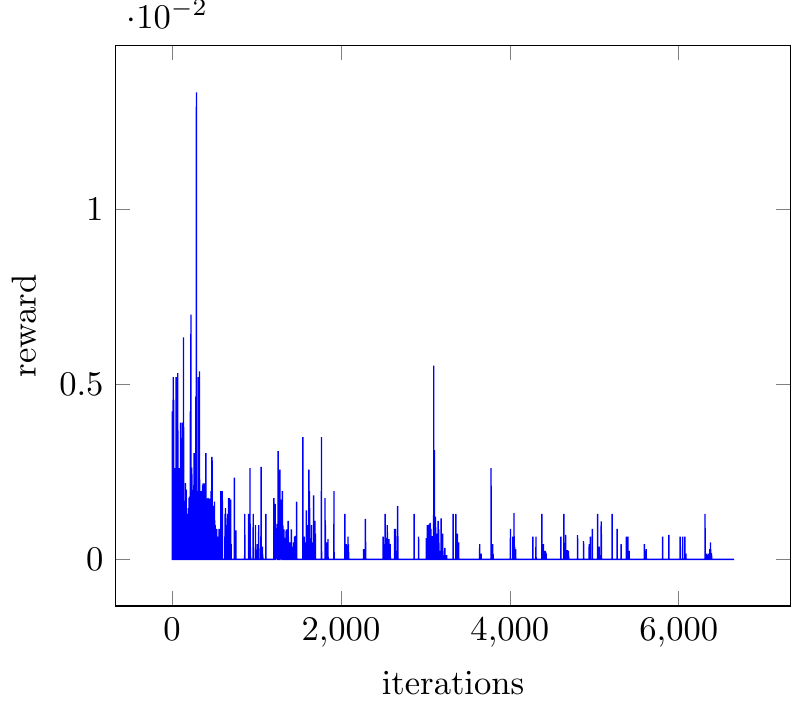}}\newline
\subfloat[objdump training reward vs. iterations]{
\includegraphics{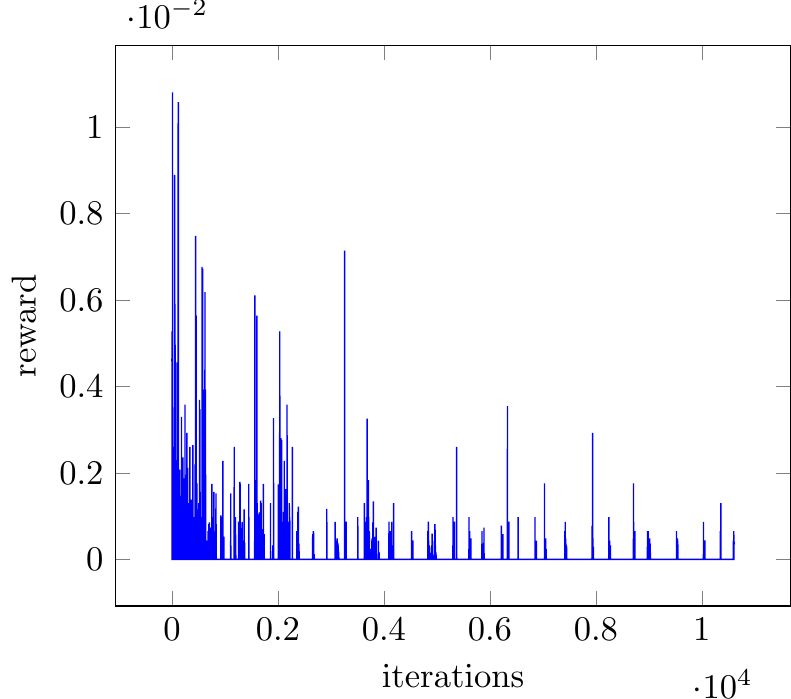}}\quad
\subfloat[readelf training reward vs. iterations]{
\includegraphics{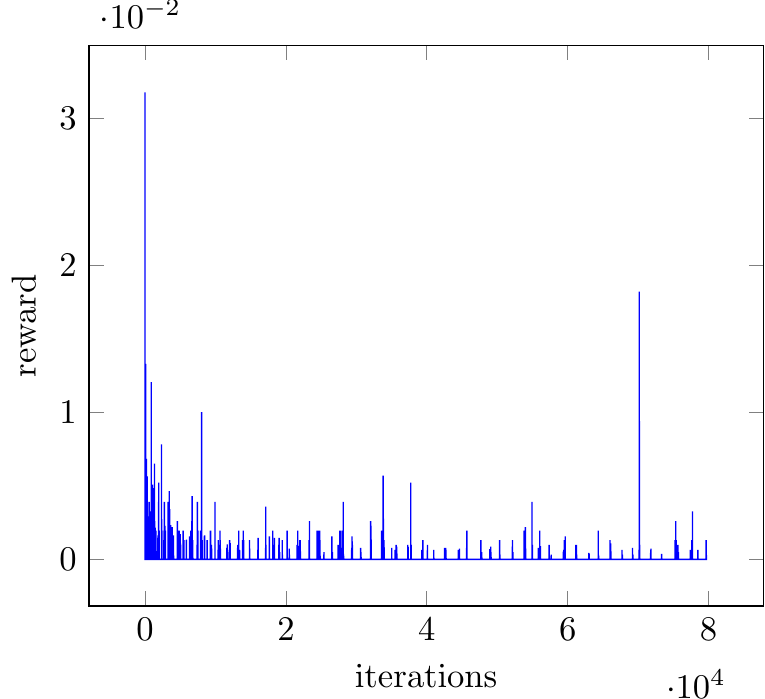}}\newline
\end{figure*}
\begin{figure*}
\subfloat[size training reward vs. iterations]{
\includegraphics{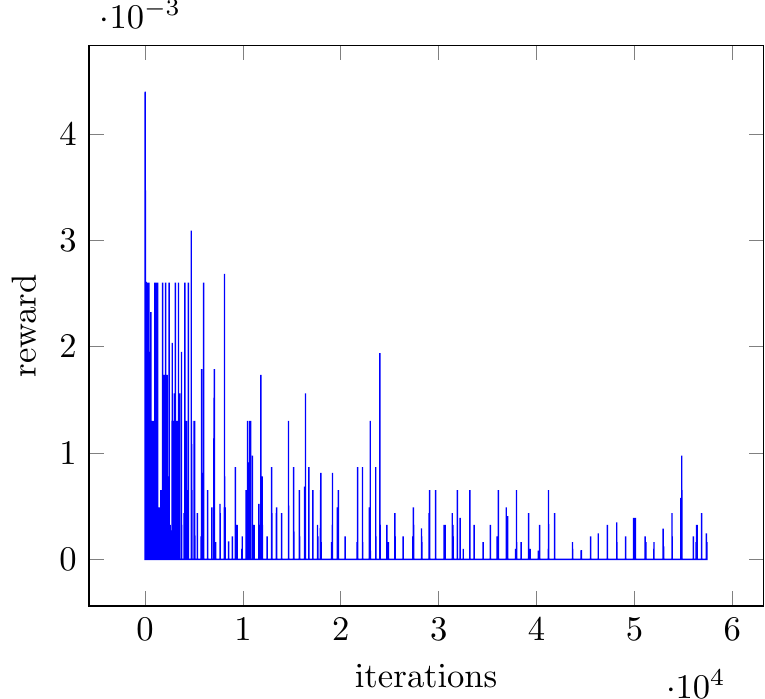}}\quad
\subfloat[strip-new training reward vs. iterations]{
\includegraphics{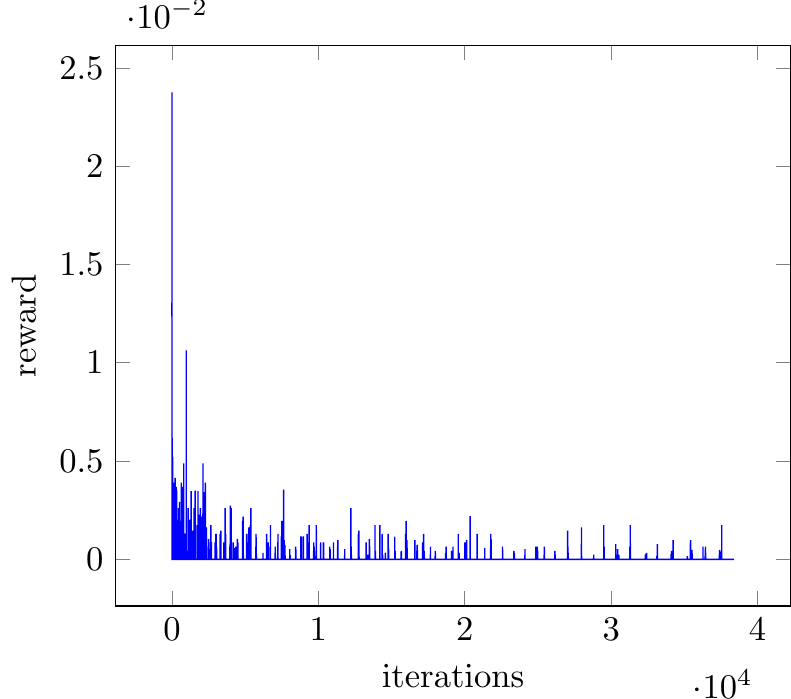}}\newline
\subfloat[gif2png training reward vs. iterations]{
\includegraphics{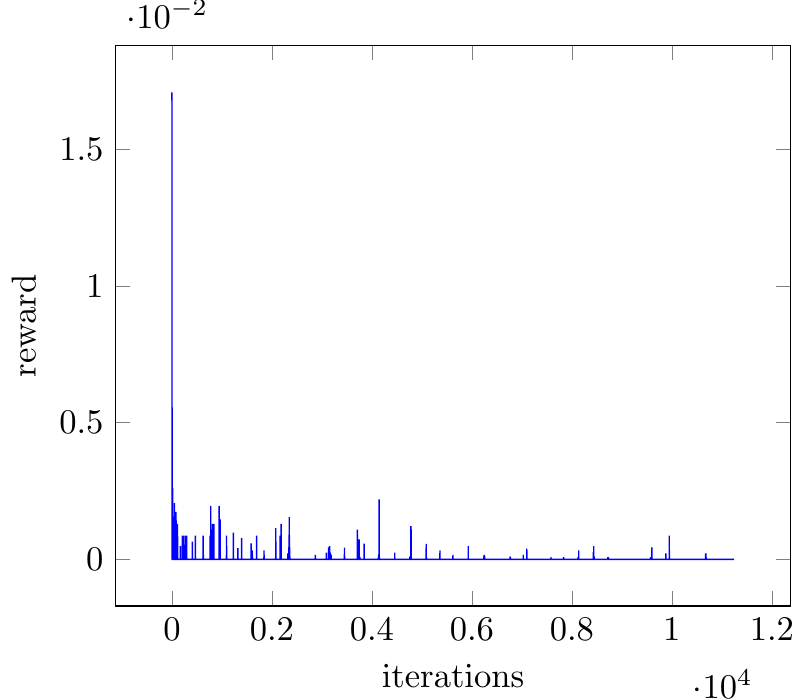}}\quad
\subfloat[libxml2 training reward vs. iterations]{
\includegraphics{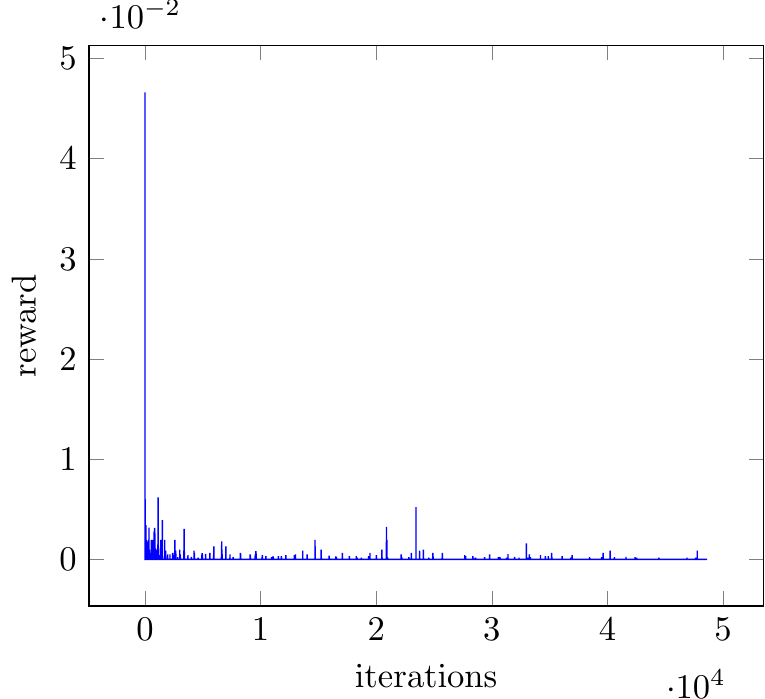}}\newline
\subfloat[libpng training reward vs. iterations]{
\includegraphics{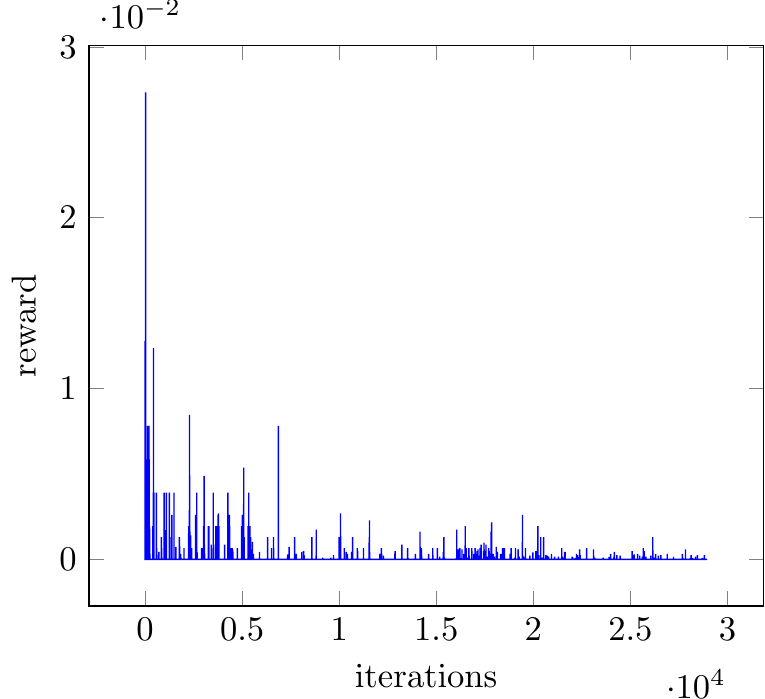}}\quad
\subfloat[tcpdump training reward vs. iterations]{
\includegraphics{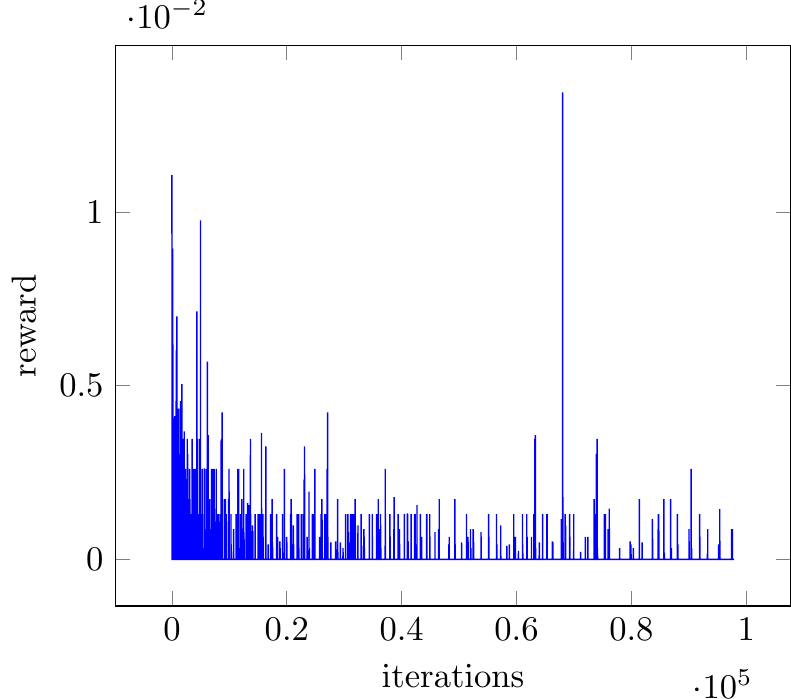}}\newline
\end{figure*}

In first experiment, we run normal AFL on the binaries addr2line -e, cxxfilt, elfedit, nm-new -C, objcopy, objdump -d, readelf -a, size, strings, strip-new, gif2png, mpg321 --stdout, pngtest, xmllint, tcpdump -nn -r for 24 hours. We run 4 instances for each binary and then take average of the coverage, total paths and crashes produced. For total paths and crashes the average values are rounded off. For each binary, we give a single seed which is provided by the AFL.
In second experiment we run AFL with `-d' option for 24 hours. `-d' option indicates afl without deterministic phase. Similar to first experiment, we run 4 instances of each binary and then average out the values. In third experiment, we train our model on each binary for 4 hours using algorithm \ref{alg:training_alg} and we save our model and reset all the things like queue, crashes, hangs etc. We then start our testing algorithm with a single input seed and run it for 24 hours. We perform this experiments on 4 instances per binary and then take average values. The results of first, second and third experiments are mentioned in the table \ref{tab:1}. 

We have also performed the cross-binary experiments across the binutils binaries, where we train our model on one binary and test it on the other binaries. The reason behind performing the cross-binary experiments is, all these binaries belong to binutils and most of them use common libraries like BFD. The coverage results are mentioned in the table \ref{tab:2}.

We have also added plots of coverage over time for all binaries for the second and third experiment in the above mentioned figures. Here we take the best instance in terms of coverage. We have also added plots of reward vs. iterations, during training phase. We found that we perform better than normal afl(i.e. AFL with deterministic phase) in terms of coverage for most of the binaries. But we perform comparatively bad than AFL without deterministic phase. We hope that the results can be further improved by careful tuning of our contextual bandit model.

\section{Related Work}

Our work is mostly inspired by the current work happening in the field of program analysis using machine/reinforcement learning. In this section we discuss these works. One of the early work in this area is Learn$\&$Fuzz\cite{godefroid2017learn}. The authors have proposed a neural network based model for format specific grammar construction. During fuzzing the AFL crates a lot of test cases which are invalid, thus Learn$\&$Fuzz paper proposed a machine learning technique to create grammar for PDF format and then they used it to create valid test cases. In our work, we focus on format independent fuzzing. Rajpal et.al.\cite{rajpal2017not} have proposed LSTM and seq-seq based model to create a function to predict the good locations in the input files to guide further mutations. This technique always takes the entire test case as an input and then decides the fuzzing locations, but the test cases can get pretty large as the time proceeds and thus querying the model may require a lot of time. Bottinger et.al. \cite{bottinger2018deep} have proposed the first work in fuzzing which involves the reinforcement learning techniques. They take the substring of bytes and represent it as state and they use deep Q-learning algorithm to choose the suitable mutation action. Most of the actions from the action space mentioned by them are PDF format specific. Also they compare their model with some baseline fuzzer and not the AFL. In our work we do not assume any baseline fuzzer. The above mentioned approaches use the machine learning techniques on fuzzing processes. A lot of work have been done in improving the fuzzing performance with the help of the conventional program analysis techniques like taint analysis, symbolic executions etc. The black box fuzzing has been made more efficient by using techniques like good quality seed selections\cite{householder2012probability, rebert2014optimizing}, proper scheduling of mutations\cite{woo2013scheduling}. Bohme et.al. \cite{bohme2017directed} gave a new tool AFL-Go which can be used for directed fuzzing and skipping the mutations in the unnecessary directions, also they have given an approach \cite{bohme2016coverage} for assigning more energy to the low frequency paths to improve the coverage. Lemieux et.al. \cite{Lemieux} targeted rare branches to rapidly increase greybox fuzz testing coverage. Some of the techniques have improved the existing fuzzers for some specific file formats.

\section{Conclusions and Future Work}
In this work, we formalize the problem of deciding the energy as a `contextual bandit problem', for the very first time. We present an algorithm to decide the energy multiplier of a test case given a fixed length contents of the test case. We implement our neural network based learning algorithm on top of the AFL and we compare results of different configurations of our model with AFL. We integrate AFL's code with a popular open-source machine learning framework tensorflow. This implementation can be useful for others while implementing any machine/reinforcement learning technique with fuzzers like AFL. We experiment our implementation on popular and large scale target programs. More careful parameter tuning and engineering work may lead the outperforming results than the existing fuzzers in term of coverage. As a future work, we can work on replacing other heuristics of AFL with some machine learning model. Some of the open problems are like deciding the next test case to be fuzzed from the queue of the test cases because the AFL selects the test cases in round robin fashion and it skips some of the test cases with some hand designed heuristics. We think that the heuristics of deciding the top$\_$rated test cases and favored test cases can also be completely removed or replaced if we are able to decide which test case is good or bad depending upon some model learned through past experiences.

    \bibliography{main}

	\end{multicols}
\end{document}